\documentclass[10pt,twocolumn,letterpaper]{article}

\usepackage[pagenumbers]{cvpr} 

\definecolor{cvprblue}{rgb}{0.21,0.49,0.74}
\usepackage[pagebackref,breaklinks,colorlinks,allcolors=cvprblue]{hyperref}

\usepackage{algorithm}
\usepackage{algorithmic}
\usepackage{amsmath}
\usepackage{amsfonts}
\usepackage{amssymb}
\usepackage{booktabs}
\usepackage{MnSymbol} 
\usepackage{multirow}
\usepackage{setspace}
\usepackage{xcolor}
\usepackage{colortbl}
\usepackage{soul}
\usepackage{graphicx}
\def\paperID{12881} 
\def\confName{CVPR}
\def\confYear{2026}

\title{Hierarchical Mixing Architecture for Low-light RAW Image Enhancement}

\author{Xianmin Chen$^{1}$ \quad Peiliang Huang$^{1,2}$ \quad Longfei Han$^{3}$\thanks{*Corresponding author} \quad Dingwen Zhang$^{4}$ \quad Junwei Han$^{5}$\\
$^{1}$USTC \\ $^{2}$Institute of Artificial Intelligence, Hefei Comprehensive National Science Center \\ $^{3}$Beijing Technology and Business University \\ $^{4}$Northwestern Polytechnical University \\ $^{5}$Chongqing University of Posts and Telecommunications\\
{\tt\small yicarlos@mail.ustc.edu.cn}}
\begin{document}
\maketitle
\begin{abstract}

With the rapid development of deep learning, low-light RAW image enhancement (LLRIE) has achieved remarkable progress. However, the challenge that how to simultaneously achieve strong enhancement quality and high efficiency still remains. Leveraging the inherent efficiency of Channel Attention and Mamba, we introduce a Hierarchical Mixing Architecture (HiMA), a hybrid LLRIE framework built upon two core modules. Specifically, we introduce Large Scale Block (LSB) for upper layers and Small Scale Block (SSB) for lower layers that reduce the parameters while improve the performance. Based on this framework, we also introduce a novel Local Distribution Adjustment (LoDA) module that adaptively aligns local feature statistics in a content-aware manner by learning to adjust regional luminance and contrast distributions. Moreover, to alleviate the domain ambiguity commonly observed in existing LLRIE pipelines, we design a Multi-Prior Fusion (MPF) module that leverages three complementary priors extracted from the first stage of the hybrid architecture to maintain domain consistency. Extensive experiments on multiple public benchmarks demonstrate that our approach outperforms state-of-the-art methods, delivering superior performance with fewer parameters. Code is available at \url{https://github.com/Cynicarlos/HiMA}.

\end{abstract}    
\section{Introduction}
\begin{figure}
    \centering
    \includegraphics[width=\columnwidth]{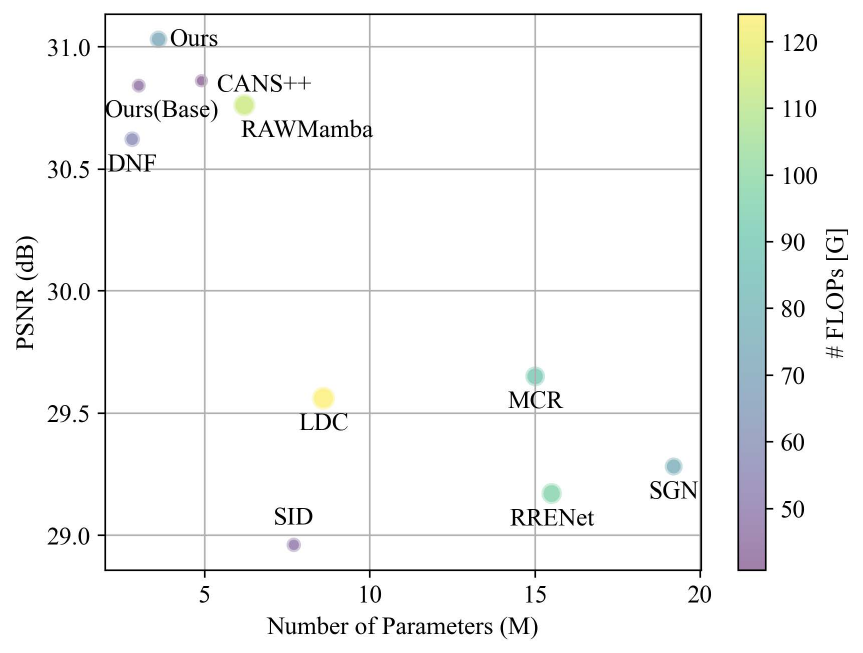}
    \captionsetup{skip=2pt} 
    \caption{Efficiency comparison. Ours method achieves superior performance in balancing number of parameters and PSNR. Even our proposed baseline can also achieve decent performance.}
    \label{fig:plot}
\end{figure}

In recent years, image processing in the RAW domain has attracted increasing attention from researchers and has demonstrated significant potential for advancing image analysis and enhancement techniques. Early studies explored the application of deep learning techniques to RAW-domain denoising. Subsequent works extended this idea by either jointly performing denoising and demosaicing within a single model or replacing the conventional ISP with deep learning under low-light conditions. Despite notable progress~\cite{DNF,MCR,cans,retinexrawmamba}, several challenges remain. In particular, designing an architecture that achieves both strong performance and high efficiency is still an open problem. This issue is especially critical for resource-constrained devices, where lightweight models with fewer parameters and reduced FLOPs are essential.

A distinctive challenge in RAW image enhancement lies in its inherently high resolution, which makes many RGB-domain restoration methods unsuitable for single-image inference without cropping. The introduction of Restormer~\cite{restormer} has significantly advanced high-resolution image restoration, while DNF~\cite{DNF} further extended its core module, Channel Self-Attention (CSA), to the RAW domain and achieved promising results. More recently, Chen et al.~\cite{retinexrawmamba} demonstrated the potential of Mamba for this task. However, their approach is highly sensitive to input resolution, making it inefficient for single-image RAW inference due to the limited GPU memory. Motivated by these observations, we introduce a Hierarchical Mixing Architecture (HiMA) that effectively balances performance and efficiency. In HiMA, Mamba is applied to the lower layers to exploit its computational advantage at reduced spatial resolutions, whereas channel attention is integrated into the upper layers to strengthen the modeling of inter-channel dependencies among large-scale features. HiMA itself achieves performance comparable to most recent state-of-the-art methods when used as a baseline, as illustrated in Fig.~\ref{fig:plot}; the full proposed method further improves upon this performance.


Another major challenge in RAW image enhancement lies in the highly uneven local exposure distributions, which are often overlooked by existing methods and lead to sub-optimal denoising performance. Most approaches adopt a global intensity scaling or a fixed exposure-time-based ratio to normalize illumination, neglecting spatial variations across the scene. Such global operations indiscriminately amplify both signal and noise, degrading image quality, especially in under- or over-exposed regions. Although some methods~\cite{retinexrawmamba,sied} introduced adaptive illumination correction modules, it still fails to fully account for local exposure characteristics. To address these issues, we propose a Local Distribution Adjustment (LoDA) module that adaptively aligns local feature statistics. By learning to adjust local luminance and contrast distributions, LoDA effectively suppresses noise amplification and preserves fine structural details, achieving more robust RAW-domain denoising under complex illumination conditions.

In addition, existing multi-stage enhancement frameworks~\cite{DNF, retinexrawmamba} typically rely on feedback mechanisms that fuse latent features between stages. Although this can facilitate information reuse, it also introduces domain ambiguity, as features from intermediate stages may not be aligned with the evolving enhancement domain. To mitigate this problem, we eliminate direct latent feature fusion and instead design a Multi-Prior Fusion (MPF) strategy. MPF uses three priors from the auxiliary branch, including the aligned RAW from LoDA, the denoised RAW from a simple denoising net, and the high frequency component from the denoised RAW, and fuses them through skip connections. This design not only preserves domain consistency but also enhances fine-detail reconstruction via high-frequency modulation guided by the learned priors.

In summary, our work presents a unified framework that jointly tackles the effectiveness, efficiency, and robustness challenges of RAW image enhancement. By integrating hierarchical feature interaction, locally adaptive exposure adjustment, and multi-prior guidance, the proposed approach achieves a balanced trade-off between number of parameters and reconstruction quality, advancing both the performance and practicality of RAW-domain enhancement.

Our main contributions can be summarized as follows:
\begin{itemize}
\item We propose a new Hierarchical Mixing Architecture (HiMA) for low-light RAW image enhancement, which integrates channel attention and Mamba at different layers for different feature scales to reduce the number of parameters while improve the performance.
\item We introduce LoDA for locally adaptive exposure correction and MPF for detail-preserving reconstruction by fusing multiple priors.
\item Extensive experiments on the SID, MCR, and ELD datasets demonstrate the effectiveness of our method, which achieves state-of-the-art results across multiple metrics with fewer or comparable parameters and FLOPs.
\end{itemize}
\section{Related Works}
\label{sec:releted works}
\subsection{Low Light RAW Image Enhancement}
Replacing traditional ISP pipelines with deep learning–based methods has achieved remarkable performance in recent years. Existing methods can be roughly categorized into two groups based on whether the input RAW image is degraded or not. The first group assumes clean RAW inputs and focuses on mapping them to the target sRGB outputs~\cite{isp0, isp1, isp2, isp3, isp4, isp5, isp6, isp7, isp8}, often with DSLR (Digital Single-Lens Reflex)-like styles. This can be regarded as a cross-domain style transfer problem, frequently referred to as pure AI ISP. However, in real-world scenarios, RAW images are often degraded due to environmental factors such as low illumination at night, where images typically suffer from severe noise.

\begin{figure*}[htbp]
    \centering
    \includegraphics[width=\textwidth]{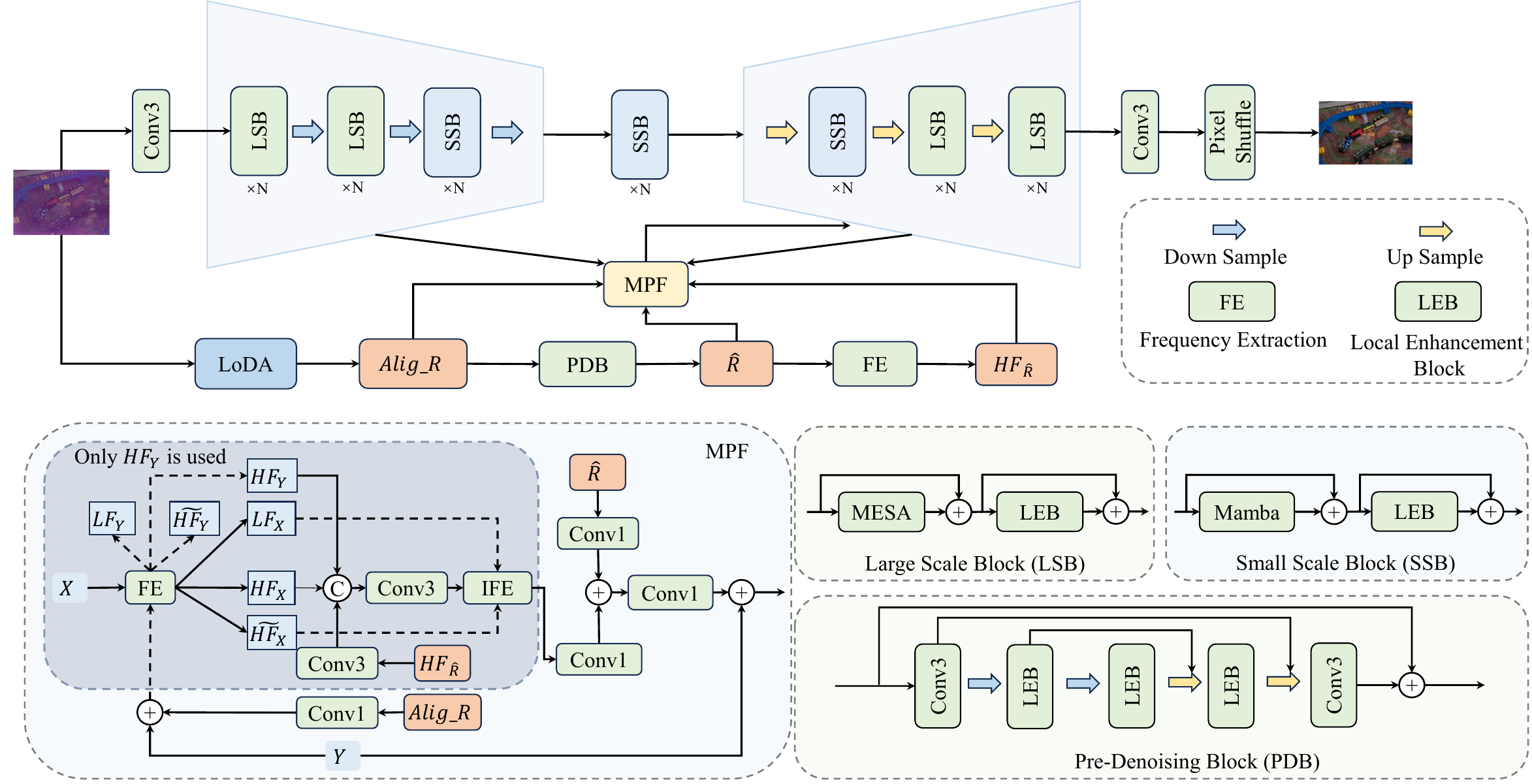}
    \captionsetup{skip=2pt} 
    \caption{Details of the overall pipeline. The original input will pass two branches, the lower one is for pre-denoising and obtaining three priors that will be used in the upper branch. While the upper branch is the main branch with HiMA, and the MPF is used for the skip connection with the priors at the same layer.}
    \label{fig:overall}
\end{figure*}

To address this problem, numerous works have been proposed in recent years~\cite{raw4, raw5, raw6, raw7, raw8, raw9}. The SID dataset and baseline model~\cite{SID} set a widely adopted benchmark for mapping noisy low-light RAW images to normally exposed sRGB images. Building upon this, subsequent methods such as DID, SGN, RRENet, LDC, and EEMEFN~\cite{DID, SGN, RRENet, LDC, EEMEFN} introduced more powerful architectures that progressively improved denoising and reconstruction quality. In recent years, multi-stage frameworks~\cite{MCR, DNF, retinexrawmamba, decouplenet} have gained popularity, as they decompose the complex RAW-to-sRGB mapping into multiple sub-tasks. For instance, DNF~\cite{DNF} introduces a decoupled and feedback strategy with a weight-shared encoder to reduce the number of parameters while maintaining good performance. However, the feedback mechanism and shared encoder can introduce domain ambiguity, resulting in unsatisfactory visual quality. To mitigate this, Chen et al.~\cite{retinexrawmamba} remove the shared encoder and introduce a Mamba-based model, which improves the scanning mechanism according to the demosaicing principle. While effective, the scanning operation over large RAW images is computationally expensive, leading to slow inference. Alternatively, CANS~\cite{cans} strengthens the shared encoder by introducing a ``backbone-head" architecture, which leverages a more expressive shared parameter space for enhanced performance.

Meanwhile, diffusion-based approaches~\cite{isp9,sied,rddm} have also emerged. Jiang et al.~\cite{sied} explored diffusion models for low-light RAW restoration, leveraging their generative ability and intrinsic denoising properties to recover visually pleasing results from extremely low-SNR RAW inputs. However, diffusion models are computationally intensive, making them impractical for single RAW image inference and deployment on resource-constrained devices.

\subsection{Mamba in Vision Tasks}
Recently, Mamba has shown promise in low-level vision tasks, such as image restoration~\cite{mambaIR, vmambaIR, freqmamba, ushaped_vision_mamba, efficientvmamba}.
VMamba \cite{vmamba} incorporates a Cross-Scan Module (CSM), which converts the input image into sequences of patches along both horizontal and vertical axes, enabling the scanning of sequences in four distinct directions. MambaIR~\cite{mambaIR} proposed a general image restoration model as a brand-new baseline. FreqMamba~\cite{freqmamba} introduces a complementary interaction structure, combining spatial Mamba, frequency band Mamba, and Fourier global modeling, to leverage Mamba's synergy with frequency analysis for tasks like image deraining. 


\textbf{Mamba–Transformer Hybrid Architecture.}
Recently, hybrid architectures combining Mamba and Transformer have been explored across various vision tasks~\cite{hybrid1,hybrid2,hybrid3,hybrid4,hybrid5,hybrid6,hybrid7}.  Zhang et al.~\cite{hybrid4} introduced HMT-UNet, the first unified framework of SSMs and Transformers for medical image segmentation. Wen et al.~\cite{hybrid2} developed a hybrid model for image restoration, integrating Mamba’s efficiency with Transformer-based contextual reasoning. Sun et al.~\cite{hybrid7} further proposed a dual-branch Transformer–Mamba network for image deraining but at the expense of increased computational cost. However, most existing designs are tailored for moderate-resolution inputs and are less effective for high-resolution data. Our work instead adopts a hierarchical integration strategy, making it more suitable for RAW-domain image enhancement.
\section{Method}
\label{sec:method}

\subsection{Overall Pipeline}
\label{sec:overall_pipeline}
As illustrated in Fig.~\ref{fig:overall}, our pipeline consists of two cooperative branches. The lower branch is designed to generate three priors that are subsequently used in the Multi-Prior Fusion (MPF) module, while the upper branch performs the main RAW-to-RGB enhancement.\\
Given a noisy low-light RAW input $X_{RAW} \in \mathbb{R}^{C_{in} \times H \times W}$, the lower branch first applies the Local Distribution Adjustment (LoDA) module to adaptively refine local feature distributions, producing the aligned RAW image $Aligned_R \in \mathbb{R}^{C_{in} \times H \times W}$. This aligned output is then fed into the Pre-Denoising Block (PDB) to obtain a preliminary denoised result $\hat{R} \in \mathbb{R}^{C_{in} \times H \times W}$. Next, a Frequency Extractor (FE) computes the high-frequency component $HF_{\hat{R}} \in \mathbb{R}^{C_{in} \times H \times W}$ from $\hat{R}$. The three priors $Aligned_R$, $\hat{R}$, and $HF_{\hat{R}}$ are then downsampled to multiple scales to facilitate hierarchical feature fusion in MPF for skip connection.\\
In the upper branch, the input $X_{RAW}$ is first processed by a shallow convolutional layer to extract low-level features, which are then fed into the Hierarchical Mixing Architecture (HiMA). Unlike traditional U-Net structures, each skip connection in HiMA is replaced with the MPF module, allowing effective integration of the three priors at corresponding scales. Finally, the aggregated features are passed through a convolutional layer followed by a pixel-shuffle operation to reconstruct the enhanced RGB image.
\begin{figure}
    \centering
    \includegraphics[width=\columnwidth]{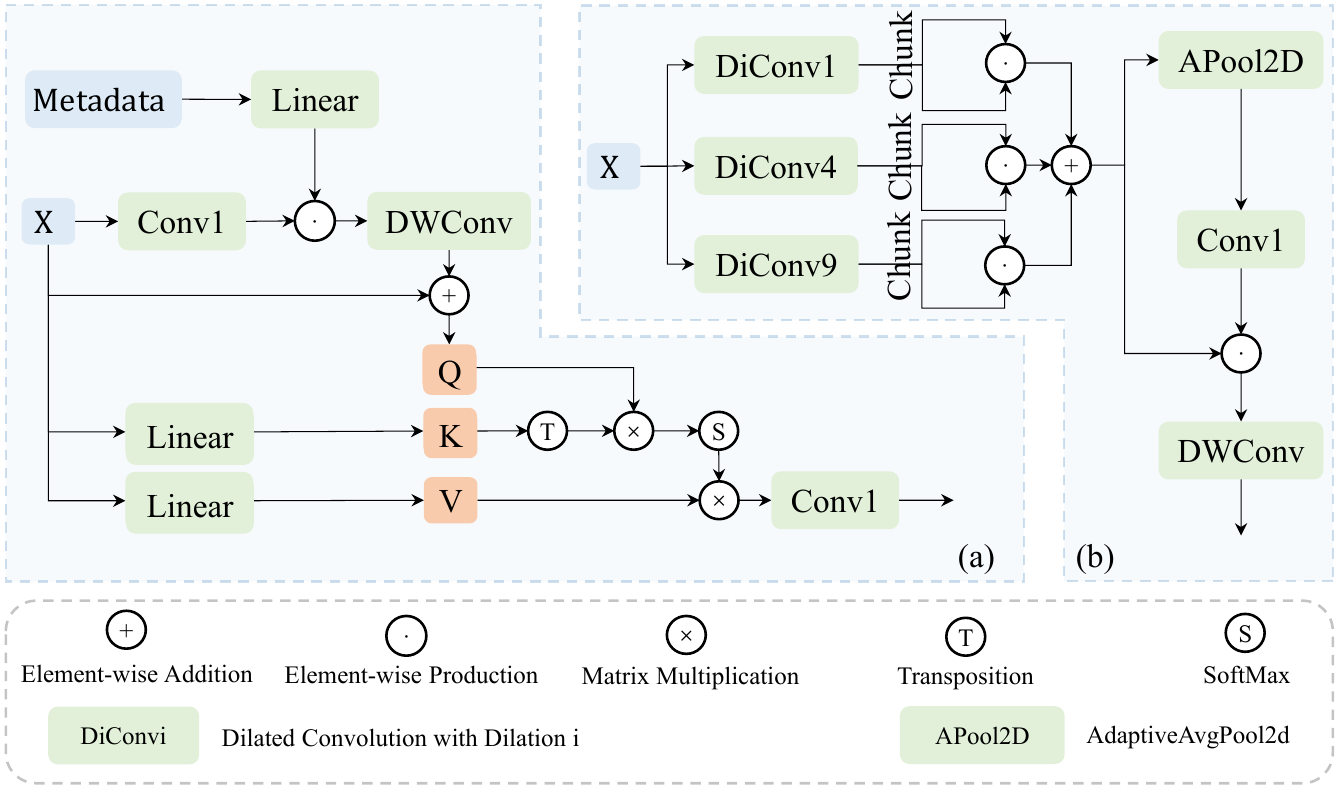}
    \captionsetup{skip=2pt} 
    \caption{Details of (a) MeSA and (b) LEB.}
    \label{fig:mesa_leb}
\end{figure}

\subsection{Hierarchical Mixing Architecture (HiMA)}
U-Shape network is widely used in low-level vision tasks, which typically consists down and up sampling to reduce the amount of computation. For low-light RAW image enhancement, the resolution of the input is usually very large, and how to efficiently perform single whole image inference becomes a big challenge. Restormer~\cite{restormer} provides a channel-wise self-attention to address this limitation for high resolution RGB images. However, it is sensitive to the number of channels, which will increase the number of parameters when the number of channels is large. The Mamba-base method~\cite{retinexrawmamba} provides an approach that the scanning is slow when the feature size is large. In addition, it's not hard to find the characteristics of U-Net are that the upper layer features have larger size and fewer channels, and the lower layer features have smaller size and more channels. Therefore, we propose HiMA, which consists Large Scale Block with channel wise attention and Small Scale Block with Mamba at different layers to make the whole model efficient.\\
\begin{figure}
    \centering
    \includegraphics[width=\columnwidth]{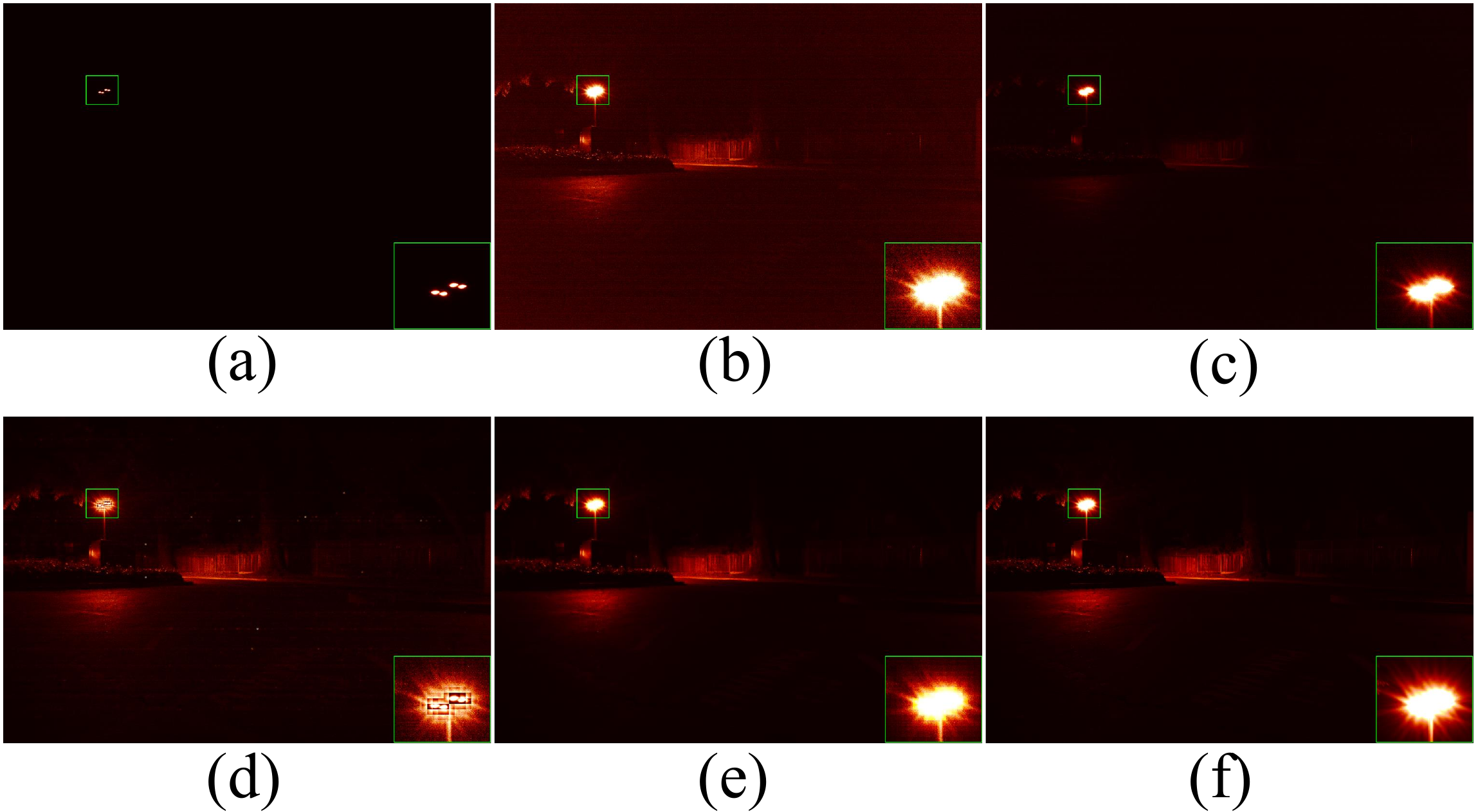}
    \captionsetup{skip=2pt} 
    \caption{Motivation of LoDA. The heatmap of the (a) Original noisy RAW, (b) Global Fixed Ratio, (c) Global Aligned Ratio, (d) Local Aligned without Std, (e) Local Aligned with Std, (f) $GT_{RAW}$. 
    }
    \label{fig:motivation2}
\end{figure}
\textbf{Large Scale Block (LSB)}
We propose LSB to deal with the large scale features at the top two layers of the U-shape network.
LSB contains two main modules, Meta Self-attention and Local Enhancement Block, and the details are shown in Fig. \ref{fig:overall}.\\
\textbf{Meta Self-attention (MeSA)}
It has been demonstrated that metadata can improve reconstruction quality in AI ISP tasks~\cite{isp7}. Therefore, we integrate metadata into channel-wise attention as MeSA. Since not all dataset has metadata, we propose use learnable metadata instead of the specific one. Specially, as shown in Fig. \ref{fig:mesa_leb} (a), for any given feature $X \in \mathbb{R}^{C\times H\times W}$ and learnable metadata $M \in \mathbb{R}^{D}$,  
\begin{equation}
    \begin{aligned}
        Q &= X + DWConv(R(Liner(M)) \cdot Conv_1(X)) \\
        K &= R(Linear(R(X))) \\
        V &= R(Linear(R(V))) \\
        Attn &= Softmax(Q \otimes K^T \cdot t) \\
        Out &= Conv_1(R(Attn \otimes V))\\
    \end{aligned}
\end{equation}
where $R(\cdot)$ is reshape function, $Conv_1$ is convolution with kernel size $1\times 1$, $\otimes$ is matrix multiplication, $t$ is a learnable scale.\\
\textbf{Small Scale Block (SSB)}
Similar to LSB, SSB contains two main modules, Mamba and Local Enhancement Block, and the details are shown in Fig.~\ref{fig:overall}. We adopt the Mamba block used in~\cite{retinexrawmamba}, but just common four-direction scanning is used instead of their eight-direction ones.\\
\textbf{Local Enhancement Block (LEB)}
We propose Local Enhancement Block as our main module in Pre-Denoising Block, and the feed forward network in both LSB and SSB. LEB uses three dilated convolutions with different dilation rates to capture features with different receptive fields. Specifically, as shown in Fig.~\ref{fig:mesa_leb} (b), given any input $X \in \mathbb{R}^{B\times C\times H\times W}$, 
\begin{equation}
    \begin{aligned}
        X_i, T_i &= DiConv_i(X).chunk(2, dim=1) \\
        X &= \sum_{i=1}^3 X_i \cdot T_i \\
        X &= Conv_1(AvgPool(X)) \cdot X \\
        X &= DWConv(X)
    \end{aligned}
\end{equation}
where $DiConv_i(\cdot)$ indicates $3\times 3$ convolution with dilation rate $i^2, i \in [1,2,3]$, $AvgPool(\cdot)$ indicates AdaptiveAvgPool2d, $Conv_1(\cdot)$ is convolution with kernel size $1 \times 1$ and $DWConv(\cdot)$ is a depthwise convolution.

\begin{algorithm}
	\setstretch{1.2}
	\renewcommand{\algorithmicrequire}{\textbf{Input:}}
	\renewcommand{\algorithmicensure}{\textbf{Output:}}
	\caption{\textbf{Lo}cal \textbf{D}istribution \textbf{A}djustment (LoDA)}
	\label{alg:loda}
	\begin{algorithmic}[1]
        \REQUIRE $x\in \mathbb{R}^{B\times C\times H\times W}, patch\_sizes$
        \STATE Initialize $aligned\_xs \leftarrow []$
        \FOR{$ps \in patch\_sizes$}
		\STATE $\mu, \sigma \leftarrow get\_local\_mean\_std(x, ps)$
        \STATE $\mu' \leftarrow Conv\_\mu_{ps}(\mu)$
        \STATE $\sigma' \leftarrow Conv\_\sigma_{ps}(\sigma)$
        \STATE $\mu' \leftarrow \mu + \mu'$
        \STATE $\sigma' \leftarrow \sigma \times exp(\sigma')$
        \STATE $x' \leftarrow \frac{x - \mu}{\sigma} \sigma' + \mu'$
        \STATE $aligned\_xs.append(x')$
        \ENDFOR
        \STATE $x \leftarrow cat(aligned\_xs, dim=1)$
        \STATE $x \leftarrow Conv(x)$
		\ENSURE $x$
	\end{algorithmic}
\end{algorithm}

\subsection{Local Distribution Adjustment (LoDA)}
As shown in Fig. \ref{fig:motivation2}, the commonly used preprocess of the noisy input RAW images is just multiplies by a fixed ratio to make the low light RAW images more close to the normal light RAW images. But we can see that the noise is also magnified with the value of the whole image is magnified. And we did experiments that took different approaches to magnify the noisy RAW images ($X_{RAW}$). We assume the distribution of the normal light RAW image ($GT_{RAW}$) is known , and we multiply $X_{RAW}$ by a variable ratio which can align the mean of $X_{RAW}$ and $GT_{RAW}$, and the result is shown in (c). The noise is less than (b) but the bright area is still not good enough, which is because the image itself is mostly dark and global magnification will affect the bright parts. Thus we consider the locality by magnifying $X_{RAW}$ locally with local size 16 $\times$ 16, we can see the whole result in (d) is better but there are obvious grids for the bright part. A question arises, if we can align the mean locally, what about the variance? Result in (e) answers with ``yes", which makes the $X_{RAW}$ more close to the $GT_{RAW}$ in (f). We can use a linear transformation to align the distribution as follows,
\begin{equation}
    \begin{aligned}
        & X' = \frac{X-\mu}{\sigma}\sigma' + \mu'
    \end{aligned}
\end{equation}
where $\mu$ and $\sigma$ is mean and standard deviation of $X_{RAW}$, respectively, $\mu'$ and $\sigma'$ is mean and standard deviation of $GT_{RAW}$, respectively.
Now, we can calculate the statistics of the transformed $X'$ as follows,
\begin{equation}
    \begin{aligned}
        E(X') &= E(\frac{X-\mu}{\sigma}\sigma' + \mu') \\
              &= \mu'
    \end{aligned}
\end{equation}
\begin{equation}
    \begin{aligned}
        Var(X') &= Var(\frac{X-\mu}{\sigma}\sigma' + \mu') \\
              &= {\sigma'}^2
    \end{aligned}
\end{equation}
Now, the local distribution is aligned to the $GT_{RAW}$.\\
However, all of these experiments have the same premise that the distribution of the $GT_{RAW}$ is known, which is not practical in real life. Therefore, we propose LoDA, which can adaptively adjust the distribution locally. And considering the effectiveness of different local size, we adapt various local size to separately adjust and fuse them finally. We also add $\epsilon$ to avoid situations where the dividend is zero. The details of LoDA are shown in  Alg.~\ref{alg:loda}.

\begin{table*}
\centering
\renewcommand\arraystretch{1.2} 
\setlength\tabcolsep{4.7pt}
\captionsetup{skip=2pt} 
\caption{Quantitative results of RAW-based LLIE methods on the Sony and Fuji subsets of SID~\cite{SID}. The top-performing result is highlighted in \textbf{bold}, the second-best is shown in \underline{underline}. Metrics marked with $\uparrow$ indicate that a higher value is better, and those marked with $\downarrow$ indicate that a lower value is better. `-' indicates the result is not available.}
\begin{tabular}{cccccccccc}
\toprule[1.2pt]
\multirow{2}{*}{Method} & \multirow{2}{*}{Venue} & \multirow{2}{*}{\#Params.(M)} & \multirow{2}{*}{\#FLOPs.(G)} & \multicolumn{3}{c}{SID Sony} & \multicolumn{3}{c}{SID Fuji} \\ \cline{5-10} \addlinespace
&  &  & & PSNR$\uparrow$ & SSIM$\uparrow$ & LPIPS$\downarrow$ & PSNR$\uparrow$ & SSIM$\uparrow$ & LPIPS$\downarrow$\\ \hline \addlinespace
SID~\cite{SID}           & CVPR2018  & 7.7   & 48.5    & 28.96 & 0.787 & 0.356  & 26.66 & 0.709 & 0.432\\
DID~\cite{DID}           & ICME2019  & 2.5   & 669.2  & 29.16 & 0.785 & 0.368  & - & - & - \\ 
SGN~\cite{SGN}           & ICCV2019  & 19.2  & 75.5   & 29.28 & 0.790 & 0.370  & 27.41 & 0.720 & 0.430\\
LLPackNet~\cite{LLPackNet}     & BMVC2020  & 1.2  &  7.2  & 27.83 & 0.755 & 0.541  & - & - & -\\
RRT~\cite{RRT}           & CVPR2021  & 0.8  & 5.2 & 28.66 & 0.790 & 0.397  & 26.94 & 0.712 & 0.446\\ 
EEMEFN~\cite{EEMEFN}     & AAAI2020  & 40.7 & 715.6  & 29.60 & 0.795 & 0.350  & 27.38 & 0.723 & 0.414\\
LDC~\cite{LDC}           & CVPR2020  & 8.6  & 124.1  & 29.56 & \underline{0.799} & 0.359  & 27.18 & 0.703 & 0.446\\
MCR~\cite{MCR}           & CVPR2022  & 15.0  & 90.5  & 29.65 & 0.797 & 0.348  & - & - & -\\ 
RRENet~\cite{RRENet}     & TIP2022   & 15.5  & 96.8  & 29.17 & 0.792 & 0.360  & 27.29 & 0.720 & 0.421\\ 
DNF~\cite{DNF}           & CVPR2023  & 2.8 & 57.0    & 30.62 & 0.797 & 0.343  & 28.71 &  0.726 & \textbf{0.391} \\ 
FDA~\cite{fda} &TIP2025 & 1.7  & -      & 30.79 & 0.797 & - & 28.88 & 0.728 & -\\
CANS++~\cite{cans} & CVPR2025  & 4.9  & 40.7  & \underline{30.86} & \underline{0.799} & \underline{0.330} & \underline{28.89} & \underline{0.728} & - \\ \hline \addlinespace 
Ours  & -          & 3.6  & 72.2 & \textbf{31.03} & \textbf{0.813} & \textbf{0.316} & \textbf{28.91} & \textbf{0.739} & \underline{0.406} \\ \bottomrule[1.2pt]
\end{tabular}
\label{table:sid}
\end{table*}

\subsection{Multi-Prior Fusion (MPF)}
We propose MPF to replace the traditional skip connection by using multi-priors. The details are shown in Fig.~\ref{fig:overall}. At each layers with same scales, five feature maps are input to MPF, which include $X, Y \in \mathbb{R}^{C\times H\times W}, \hat{R}, HF_{\hat{R}}, Aligned_R \in \mathbb{R}^{C_{in}\times H\times W}$. X and Y are the decoder's and encoder's features of HiMA at the each same layer, respectively, and the other three priors are from the output of the lower branch as introduced in Sec.~\ref{sec:overall_pipeline}. Specifically, we firstly apply a convolution to $Aligned_R$ and fuse it with $Y$ by addition. Then we apply a Frequency Extraction (FE, see Supplementary Material) to both X and the fused Y. The obtained $LF_X, HF_X, \widetilde{HF_X}, HF_Y$ are used for further 
feature fusion. Only the high-frequency components are modulated by concatenation and convolution and the $LF_X, \widetilde{HF_X}$ are unchanged for Inverse of Frequency Extraction (IFE, see Supplementary Material). After the modulation in frequency domain, the obtained feature and the prior $hat{R}$ are then fused by convolutions and additions. Finally, another convolution and a residual connection with $Y$ by addition are applied to get the output of MPF. 

\begin{table}
\vspace{0.5cm}
\centering
\renewcommand\arraystretch{1.2} 
\captionsetup{skip=2pt} 
\caption{Quantitative results on MCR~\cite{MCR} dataset. The top-performing result is highlighted in \textbf{bold}, the second-best is shown in \underline{underline}. Metrics marked with $\uparrow$ indicate that a higher value is better.}
\setlength{\tabcolsep}{7mm}{
\begin{tabular}{cccc}
\toprule[1.2pt]
Method      & PSNR $\uparrow$             & SSIM $\uparrow$   \\ \hline \addlinespace
SID~\cite{SID}        & 29.00                       & 0.906             \\ 
DID~\cite{DID}        & 26.16                       & 0.888             \\ 
SGN~\cite{SGN}        & 26.29                       & 0.882             \\ 
LDC~\cite{LDC}        & 29.36                       & 0.904             \\ 
MCR~\cite{MCR}        & 31.69                       & 0.908             \\ 
DNF~\cite{DNF}        & 32.00     & 0.915          \\ 
FDA~\cite{fda}        & 32.34     & \underline{0.917}          \\ 
CANS++~\cite{cans}   & \underline{33.09}     & 0.916          \\
\textbf{Ours}           & \textbf{34.40}      & \textbf{0.924}          \\ \bottomrule[1.2pt]
\end{tabular}}
\label{table:mcr}
\end{table}

\section{Experiments}
\label{sec:exp}

\begin{table*}
  \centering
  \renewcommand\arraystretch{1.2} 
  \captionsetup{skip=5pt} 
  \caption{Quantitative results on ELD~\cite{raw8} dataset. Several pretrained models on SID Sony dataset are directly applied to ELD dataset, including SonyA7S2 and NikonD850 subsets. The top-performing result is highlighted in \textbf{bold}, while the second-best is shown in \underline{underline}. Higher ratio represents more noise. Metrics marked with $\uparrow$ indicate that a higher value is better.}
  \setlength{\tabcolsep}{1.2mm}{
\begin{tabular}{c|cccccc|cccccc}
\hline
Dataset & \multicolumn{6}{c|}{SonyA7S2}                 & \multicolumn{6}{c}{NikonD850}                 \\
\hline
Ratio & \multicolumn{2}{c}{100} & \multicolumn{2}{c}{200} & \multicolumn{2}{c|}{Avg} & \multicolumn{2}{c}{100} & \multicolumn{2}{c}{200} & \multicolumn{2}{c}{Avg} \\
\hline
Method  & PSNR$\uparrow$  & SSIM$\uparrow$  & PSNR$\uparrow$  & SSIM$\uparrow$  & PSNR$\uparrow$  & SSIM$\uparrow$  & PSNR$\uparrow$  & SSIM$\uparrow$  & PSNR$\uparrow$  & SSIM$\uparrow$  & PSNR$\uparrow$  & SSIM$\uparrow$  \\
\hline
SID~\cite{SID}     & 26.80 & 0.826 & 25.98 & 0.780 & 26.39 & 0.803 & 25.77 & 0.764 & 25.01 & 0.739 & 25.39 & 0.751 \\
MCR~\cite{MCR}     & 26.62 & 0.837 & 24.00 & 0.717 & 25.31 & 0.777 & 25.60 & 0.785 & 25.41 & 0.758 & 25.50 & 0.772 \\
DNF~\cite{DNF}     & \underline{28.35} & 0.869 & 26.94 & \underline{0.819} & 27.64 & \underline{0.844} & 27.51 & \underline{0.816} & 25.93 & \textbf{0.769} & 26.72 & \textbf{0.792} \\
CANS++~\cite{cans}  & 28.20 & \textbf{0.872} & \textbf{27.41} & \textbf{0.821} & \underline{27.80} & {\textbf{0.847}} & \underline{28.09} & 0.798 & \underline{26.61} & 0.746 & \underline{27.35} & 0.772 \\
Ours    & \textbf{28.55} & \underline{0.871} & \underline{27.27} & 0.813 & \textbf{27.91} & 0.842 & {\textbf{28.25}} & \textbf{0.822} & \textbf{26.80} & \underline{0.760} & \textbf{27.53} & \underline{0.791} \\
\hline
\end{tabular}}
\label{tab:eld}
\end{table*}

\subsection{Datasets}
\textbf{SID Dataset.} The Sony subset contains 1865 training pairs of short- and long-exposure RAW images ($2848\times4256$), where the short exposure serves as input and the long one as $GT_{raw}$. Following DNF~\cite{DNF}, three misaligned test scenes are removed. The Fuji subset includes 1655 training and 524 test pairs ($4032\times6032$) with an X-Trans CFA.\\
\textbf{MCR Dataset.} The MCR~\cite{MCR} dataset provides 4980 images ($1280\times1024$), including 3984 low-light RAW and 498 sRGB images under indoor (1/256s–3/8s) and outdoor (1/4096s–1/32s) exposures. RAW ground truth is obtained following DNF~\cite{DNF}, and preprocessing is similar to SID but without random cropping.\\
\textbf{ELD Dataset.} The ELD~\cite{raw8} dataset includes 10 indoor scenes captured by four cameras (SonyA7S2, NikonD850, CanonEOS70D, CanonEOS700D). We use SonyA7S2 and NikonD850 data at three ISO levels (800, 1600, 3200) and two illumination factors (100, 200), resulting in 120 RAW pairs. The SID-Sony pretrained model is used for generalization evaluation.

\subsection{Implementation Details}
All of the experiments in this paper were conducted on a single NVIDIA 3090. For SID Sony, the initial learning rate was set to 2e-4, and was reduced to 2e-5 at the 300th epochs by cosine annealing strategy. While for SID Fuji dataset, we set the initial learning rate to 1e-4, and reduced it to 1e-5 at the 100th epochs by cosine annealing strategy. For both Sony and Fuji dataset, the input was randomly cropped to $512\times512\times4$ and $512\times512\times9$, respectively. For MCR dataset, the learning rate schedule is same as that of SID Sony dataset, but the input was not cropped with shape $512\times640\times4$. And random data augmentations including horizontal/vertical flipping and transposition are used for all datasets. The batch size was set to 1 and Adamw optimizer with betas parameters [0.9,0.999] and momentum 0.9 was used, and Muon~\cite{muon} optimizer was jointly used to accelerate the convergence of training. The number of blocks at each layer of HiMA in Fig. \ref{fig:overall} $N$ is set to 2.

\subsection{Comparison with SOTA Methods}
We compared our method with most of the previous SOTA methods, including SID~\cite{SID}, DID~\cite{DID}, SGN~\cite{SGN}, LLPackNet~\cite{LLPackNet}, RRT~\cite{RRT}, EEMEFN~\cite{EEMEFN}, LDC~\cite{LDC}, MCR~\cite{MCR}, RRENet~\cite{RRENet}, DNF~\cite{DNF}, FDA~\cite{fda}, RawMamba~\cite{retinexrawmamba}, CANS++~\cite{cans}. And the quantitative evaluation metrics we used were as same as that in previous works~\cite{DNF, retinexrawmamba}, including PSNR, SSIM~\cite{ssim} and LPIPS~\cite{lpips}. 

\noindent \textbf{Quantitative Comparisons.} The quantitative results are reported in Tab.~\ref{table:sid} and Tab.~\ref{table:mcr}. Our method surpasses all previous approaches on most evaluation metrics. Specifically, on the SID Sony dataset, we achieve a 0.17 dB higher PSNR than the second-best method, CANS++, while using 1.3M fewer parameters. On the SID Fuji dataset, although the PSNR improvement is relatively modest, our method attains a 0.011 higher SSIM than CANS++. For the MCR dataset, our approach delivers substantial gains, outperforming CANS++ by 1.31 dB in PSNR and 0.008 in SSIM. In other words, our method achieves a 3.96\% PSNR increase while requiring only 73.5\% of the parameters used by CANS++.

\begin{figure*}[ht]
    \centering
    \includegraphics[width=\textwidth]{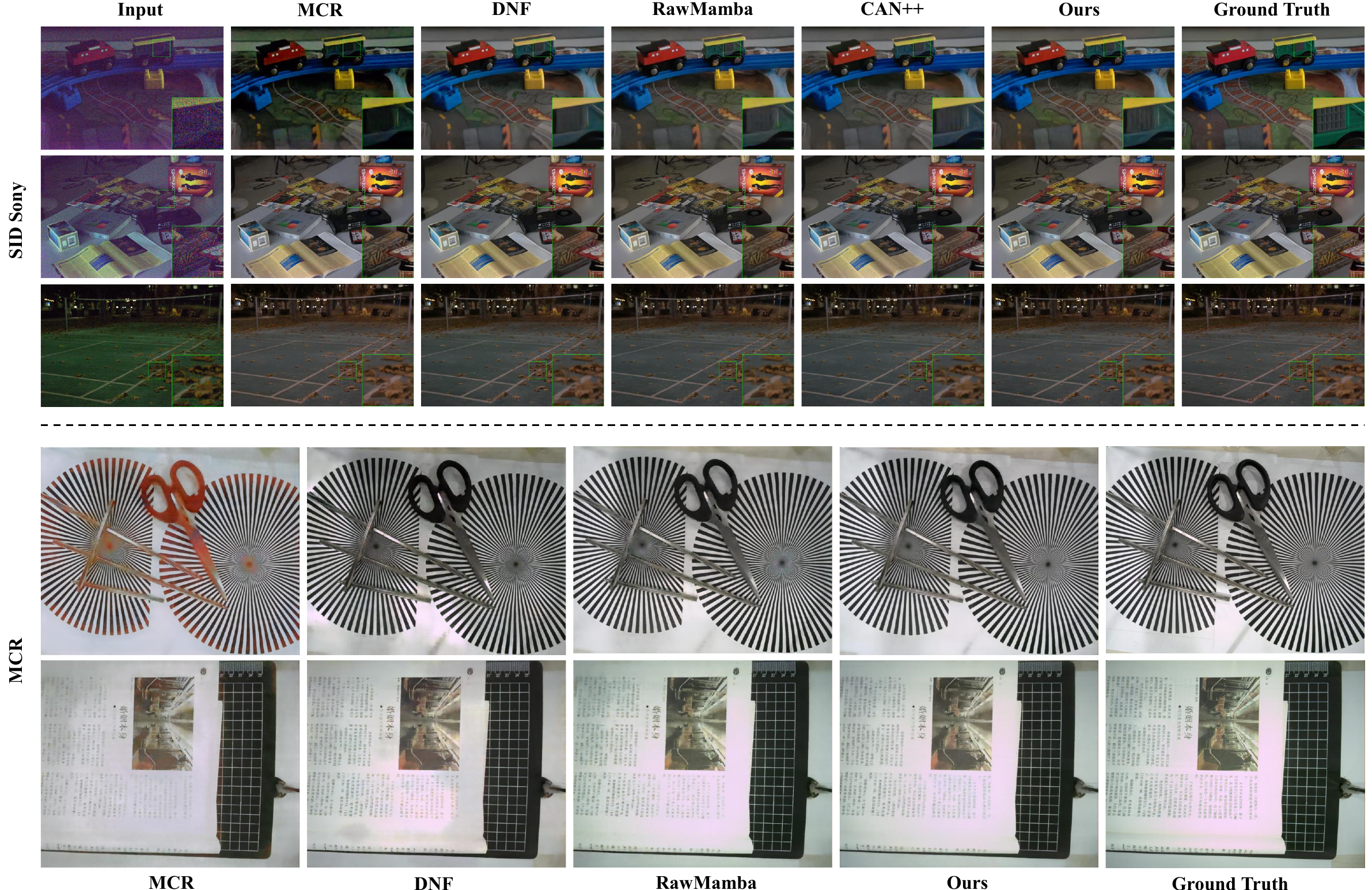}
    \caption{The visualization results between our method and the state-of-the-art methods on SID and MCR dataset. (Zoom-in for best view). 
    Note that CANS++ is not compared on MCR dataset since it did not release the pretrained model.}
    \label{fig:visual}
\end{figure*}

To further validate the generalization capability of our model, we evaluated it alongside several prior methods on the ELD dataset, using pretrained weights from the SID Sony dataset. Two subsets of ELD, namely Sony and Nikon, were tested, and the results are summarized in Tab.~\ref{tab:eld}. On the Sony subset, our method achieves performance comparable to CANS++ in both PSNR and SSIM, while on the Nikon subset, it surpasses CANS++ in both metrics and achieves SSIM comparable to DNF. Specifically, on average, our approach yields a 0.11 dB higher PSNR on the ELD Sony subset and a 0.18 dB higher PSNR on the ELD Nikon subset compared to the second-best method, demonstrating its strong generalization capability across different camera domains.

\noindent \textbf{Qualitative Comparisons.} Representative visual comparisons are shown in Fig.~\ref{fig:visual}, using examples from the SID Sony and MCR datasets. We compare our method against several recent approaches, including MCR~\cite{MCR}, DNF~\cite{DNF}, RawMamba~\cite{retinexrawmamba}, and CANS++~\cite{cans}. As illustrated, our method produces colors that are more faithful to the ground truth. For example, in the first and last scenes of the Sony dataset, the regions of interest (ROIs) magnified in the bottom-right corners show that our reconstructions preserve hue more accurately than competing methods. Also, in the first scene of the MCR dataset, our method demonstrates superior accuracy in brightness adjustment compared to other approaches, particularly in reflective regions, and exhibits minimal instances of incorrect brightness compensation. More visualization comparisons can be found in the supplementary material.

\subsection{Ablation Studies}
\noindent \textbf{Ablation on HiMA.}
To evaluate the effectiveness of HiMA, we first replaced all Small Scale Blocks (SSB) with Large Scale Blocks (LSB) while keeping all other settings unchanged, that is removing Mamba from the architecture. The results are presented in Tab.~\ref{tab:ablation_hima}. As shown, replacing SSB with LSB increases both the number of parameters and FLOPs, while reducing performance by 0.38 dB in PSNR and 0.006 in SSIM. In other words, HiMA achieves higher PSNR and SSIM with fewer parameters and FLOPs compared to a traditional architecture that uses identical blocks at all layers.
Furthermore, to test the role of Mamba within SSB, we replaced Mamba with the spatial attention module from CBAM~\cite{cbam}. As shown in the second row of Tab.~\ref{tab:ablation_hima}, although simple spatial attention reduces parameters and FLOPs slightly, it yields worse PSNR and SSIM compared to our proposed HiMA.
\begin{table}[t]
  \vspace{0.5cm}
  \centering
  \renewcommand\arraystretch{1.2} 
  \captionsetup{skip=2pt} 
  \caption{Ablation study on HiMA}
  \setlength{\tabcolsep}{2mm}{
    \begin{tabular}{c|cccc}
    \bottomrule
          & PSNR  & SSIM  &\#Params.(M) & FLOPs(G)\\
    \hline
    All\_LSB & 30.65    & 0.807    & 4.2     & 75.8\\
    SSB\_SA & 30.69     & 0.808    & 2.9     & 66.0 \\
    HiMA  & 31.03      & 0.813    & 3.6     & 72.2\\
    \toprule
    \end{tabular}}%
  \label{tab:ablation_hima}%
  \vspace{-0.5cm}
\end{table}%

\begin{table}[h]
\vspace{0.5cm}
  \centering
  \renewcommand\arraystretch{1.2} 
  \captionsetup{skip=2pt} 
  \caption{Ablation study on main modules}
  \setlength{\tabcolsep}{2.4mm}{
    \begin{tabular}{c|cccc}
    \bottomrule
    Baseline (HiMA) &$\checkmark$      &       &       &\\
    MeSA          &       &$\checkmark$       &$\checkmark$       &$\checkmark$  \\
    LoDA          &       &       &$\checkmark$       &$\checkmark$  \\
    MPF           &       &       &       & $\checkmark$ \\
    \hline
    PSNR   & 30.84      & 30.88   & 30.93       &\textbf{31.03}  \\
    SSIM   & 0.803      & 0.805   & 0.809      &\textbf{0.813}  \\
    \#Params(M)  & 3.0  & 3.1      & 3.6      & 3.6 \\
    \#FLOPs(G)   & 45.30 & 50.50      & 70.5      & 72.2 \\
    \toprule
    \end{tabular}}%
  \label{tab:ablation_main}%
\end{table}%
\noindent \textbf{Ablation study on main modules.}
We first established a simple yet effective baseline using the HiMA architecture by replacing the MPF module with a standard convolution-based fusion and removing the three priors ($AR$, $\hat{R}$, and $HF_{\hat{R}}$) as well as the learnable metadata in MeSA. All ablation experiments were performed on the SID Sony dataset, and the results are summarized in Tab.~\ref{tab:ablation_main}. Even without these additional components, the HiMA-based baseline achieves competitive results, which is only 0.02 dB lower in PSNR and with comparable SSIM to CANS++ while using merely 61.2\% of its parameters and maintaining similar computational complexity.
When learnable metadata is incorporated into all LSB blocks, the model exhibits a slight but consistent improvement, with PSNR and SSIM increasing by 0.04 dB and 0.002, respectively, at the cost of only a marginal increase in model size and FLOPs. Introducing the LoDA module further enhances the network’s denoising capability, providing an additional gain of 0.07 dB in PSNR and 0.006 in SSIM, while still preserving computational efficiency. Finally, replacing the standard convolutional fusion with the proposed MPF module, which integrates the three complementary priors, leads to the full HiMA model that achieves the best overall performance across all metrics.

\noindent \textbf{Ablation study on MPF.}
We further conducted a detailed analysis of the MPF module, with results presented in Tab.~\ref{tab:ablation_mpf}. When the high-frequency prior $HF_{\hat{R}}$ is removed, the model performance drops noticeably, with a 0.04 dB decrease in PSNR and a 0.003 reduction in SSIM, indicating the crucial role of high-frequency information in restoring fine details. In contrast, removing the reconstructed prior $\hat{R}$ causes only a marginal decline in both metrics, possibly because its feature representation overlaps with that of $AR$ and $HF_{\hat{R}}$. Finally, eliminating all priors leads to a more evident degradation, 0.05 dB in PSNR, confirming that the joint use of multiple priors contributes substantially to accurate detail reconstruction and overall enhancement quality.
\begin{table}
 \vspace{0.5cm}
  \centering
  \renewcommand\arraystretch{1.2} 
  \captionsetup{skip=2pt} 
  \caption{Ablation study on MPF}
  \setlength{\tabcolsep}{4.2mm}{
    \begin{tabular}{c|cccc}
    \bottomrule
    $Alig\_R$    & $\checkmark$     & $\checkmark$     & $\checkmark$ & \\
    $\hat{R}$     & $\checkmark$      & $\checkmark$     &  & \\
    $HF_{\hat{R}}$   & $\checkmark$      &       &  & \\
    \hline
    PSNR  & \textbf{31.03}      & 30.99      & 30.98 & 30.93\\
    SSIM  & \textbf{0.813}      & 0.810      & 0.809  & 0.809\\
    \toprule
    \end{tabular}}%
  \label{tab:ablation_mpf}%
  \vspace{-0.5cm}
\end{table}%
\section{Conclusion}
\label{sec:conclusion}

In this paper, we propose a novel and efficient hierarchical mixing architecture (HiMA) for low-light RAW image enhancement. Our method integrates channel-wise attention and Mamba modules across different feature scales, leveraging their complementary strengths to enhance performance while reducing model complexity. To overcome the limitation of fixed-ratio preprocessing, we introduce LoDA, which adaptively aligns illumination through multi-scale local distribution adjustment. Furthermore, to fully exploit the information available in the RAW domain, we design MPF to utilize high-frequency features for more accurate detail restoration during fusion. Extensive experiments on multiple public datasets, both qualitative and quantitative, demonstrate that our approach achieves superior performance with only 73.5\% of the parameters of the latest state-of-the-art method.
Although the training stability on the SID Fuji dataset remains slightly lower and deploying Mamba-based models on edge devices is still challenging, these aspects open up valuable directions for future research.
\clearpage
{
    \small
    \bibliographystyle{ieeenat_fullname}
    \bibliography{main}
}

\clearpage
\setcounter{page}{1}
\maketitlesupplementary
\section{A1. Algorithm details}
\label{sec:appendix_alg}


\begin{algorithm}
	\setstretch{1.2}
	\renewcommand{\algorithmicrequire}{\textbf{Input:}}
	\renewcommand{\algorithmicensure}{\textbf{Output:}}
	\caption{Frequency Extract (FE)}
	\label{alg2}
	\begin{algorithmic}[1]
        \REQUIRE $x \in \mathbb{R}^{B\times C\times H\times W}$
		\STATE Initialization: $threshold \leftarrow 0.01$
		\STATE  $fft\_x \leftarrow fft2(x, dim=(-2, -1))$
        \STATE $fft\_x\leftarrow fftshift(fft\_x, dim=(-2, -1))$
        \STATE $h' \leftarrow int(h \times threshold)$
        \STATE $w' \leftarrow int(w \times threshold)$
        \STATE $m\_low \leftarrow torch.zeros\_like(fft\_x)$
        \STATE $m\_low[..., \frac{h}{2} - h':\frac{h}{2} + h', \frac{w}{2} - w':\frac{w}{2} + w'] \leftarrow 1$
        \STATE $m\_high \leftarrow 1 - m\_low$
        \STATE $x\_low \leftarrow fft\_x \times m\_low$
        \STATE  $x\_high \leftarrow fft\_x \times m\_high$
        \STATE  $x\_high\_real \leftarrow torch.real(x\_high)$
        \STATE  $x\_high\_imag \leftarrow torch.imag(x\_high)$
		\ENSURE  $x\_high\_real, x\_high\_imag, x\_low$
	\end{algorithmic}  
\end{algorithm}

\begin{algorithm}
	\setstretch{1.2}
	\renewcommand{\algorithmicrequire}{\textbf{Input:}}
	\renewcommand{\algorithmicensure}{\textbf{Output:}}
	\caption{Inverse of Frequency Extract (IFE)}
	\label{alg3}
	\begin{algorithmic}[1]
        \REQUIRE $x\_high\_real, x\_high\_imag, x\_low \in \mathbb{R}^{B\times C\times H\times W}$
		\STATE $x\_high \leftarrow complex(x\_high\_real, x\_high\_imag)$
        \STATE $fft\_x \leftarrow x\_low + x\_high$
        \STATE $fft\_x \leftarrow ifftshift(fft\_x, dim=(-2, -1))$
        \STATE $x \leftarrow ifft2(fft\_x, dim=(-2, -1)).real$
		\ENSURE $x$
	\end{algorithmic}  
\end{algorithm}

\section{Loss Function}
We followed most of the previous works~\cite{DNF, retinexrawmamba} and just used L1 loss. And our method involves both RAW and sRGB domains, so the loss can be expressed as follows: 
\begin{align}
    L_{total}  = \alpha {||\hat{Y}_{raw} - GT_{raw}||}_1 + \beta {||\hat{Y}_{srgb} - GT_{srgb}||}_1 \notag
\end{align}
where $Y_{raw}$ is the raw image after denoised, e.g. $\hat{R}$ in Fig. \ref{fig:overall}, $Y_{srgb}$ is the final sRGB image that our model output, $GT_{srgb}$ is the sRGB image obtained from raw ground truth after post-processing by Rawpy as previous works did. And $\alpha$ and $\beta$ defaults to 1.0 in our experiments. Note that the RAW domain supervision is not used for our baseline with HiMA in Sec. \ref{sec:exp} since it does not use any priors.

\newpage
\section{A2. More Visual Results}
\begin{figure*}
    \centering
    \includegraphics[width=\textwidth]{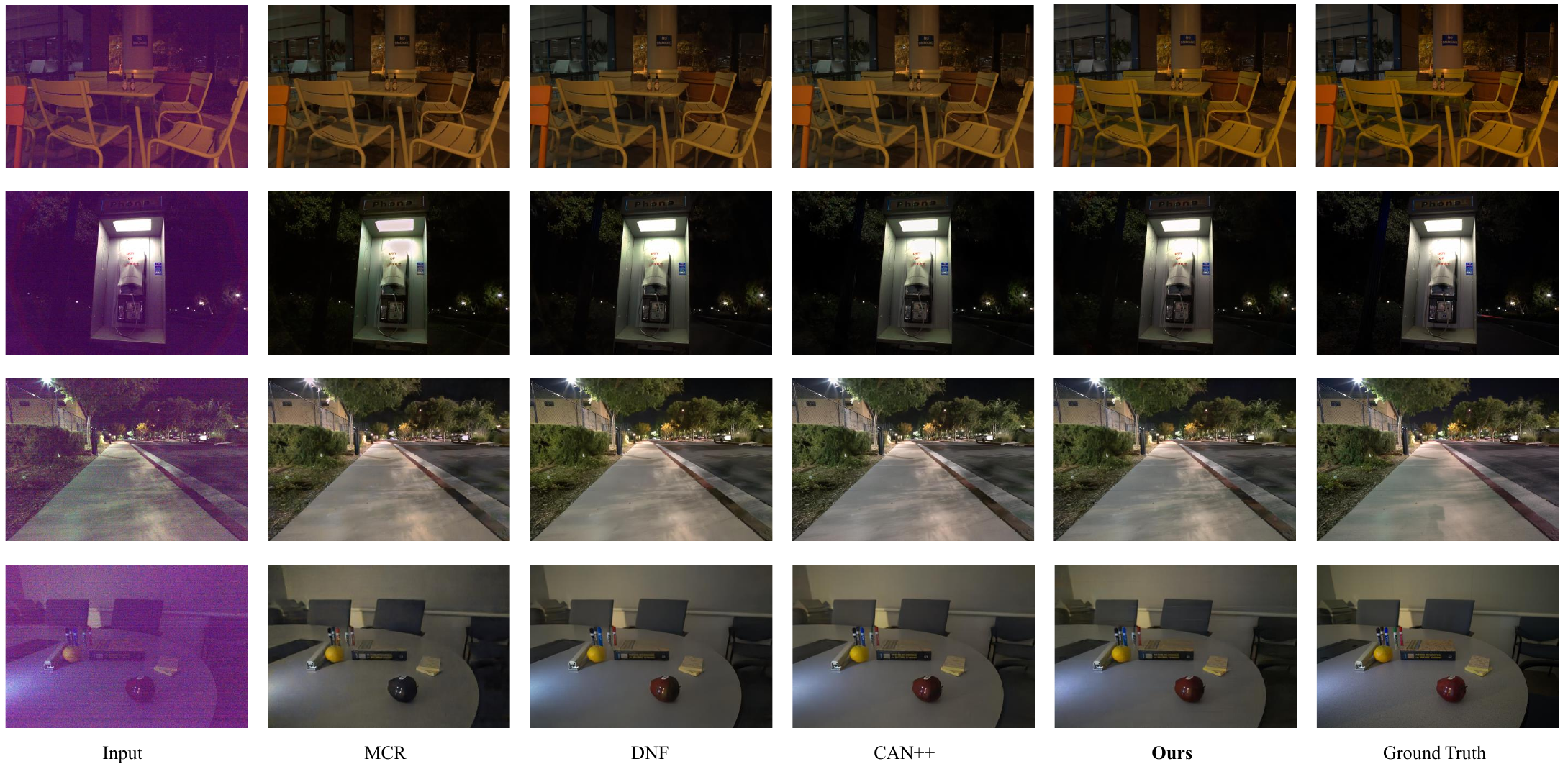}
    \caption{More visualization results between our method and the state-of-the-art methods on SID dataset. (Zoom-in for best view).}
\end{figure*}
\clearpage
\begin{figure*}
    \centering
    \includegraphics[width=\textwidth]{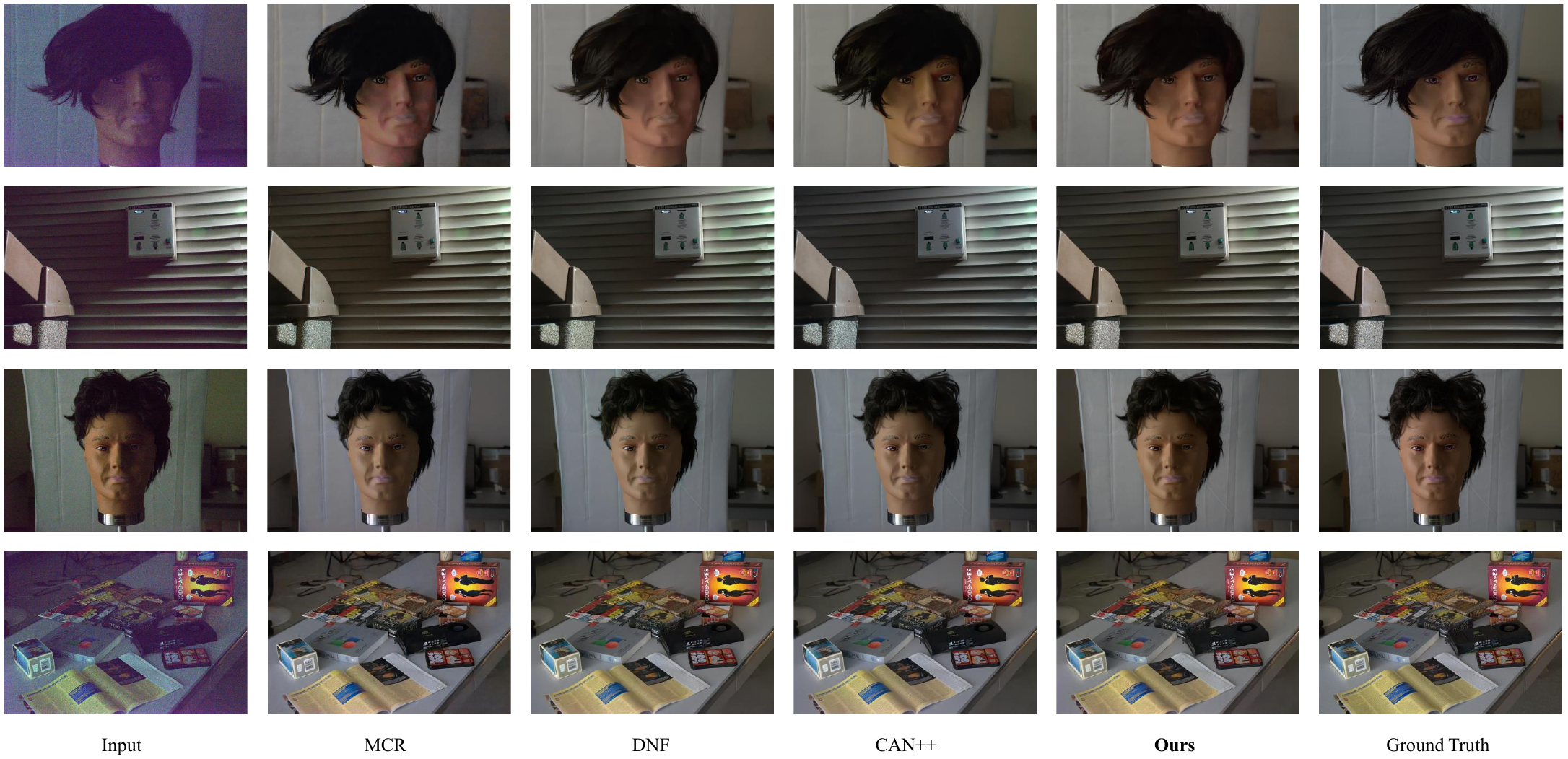}
    \caption{More visualization results between our method and the state-of-the-art methods on SID dataset. (Zoom-in for best view).}
\end{figure*}
\clearpage
\begin{figure*}
    \centering
    \includegraphics[width=\textwidth]{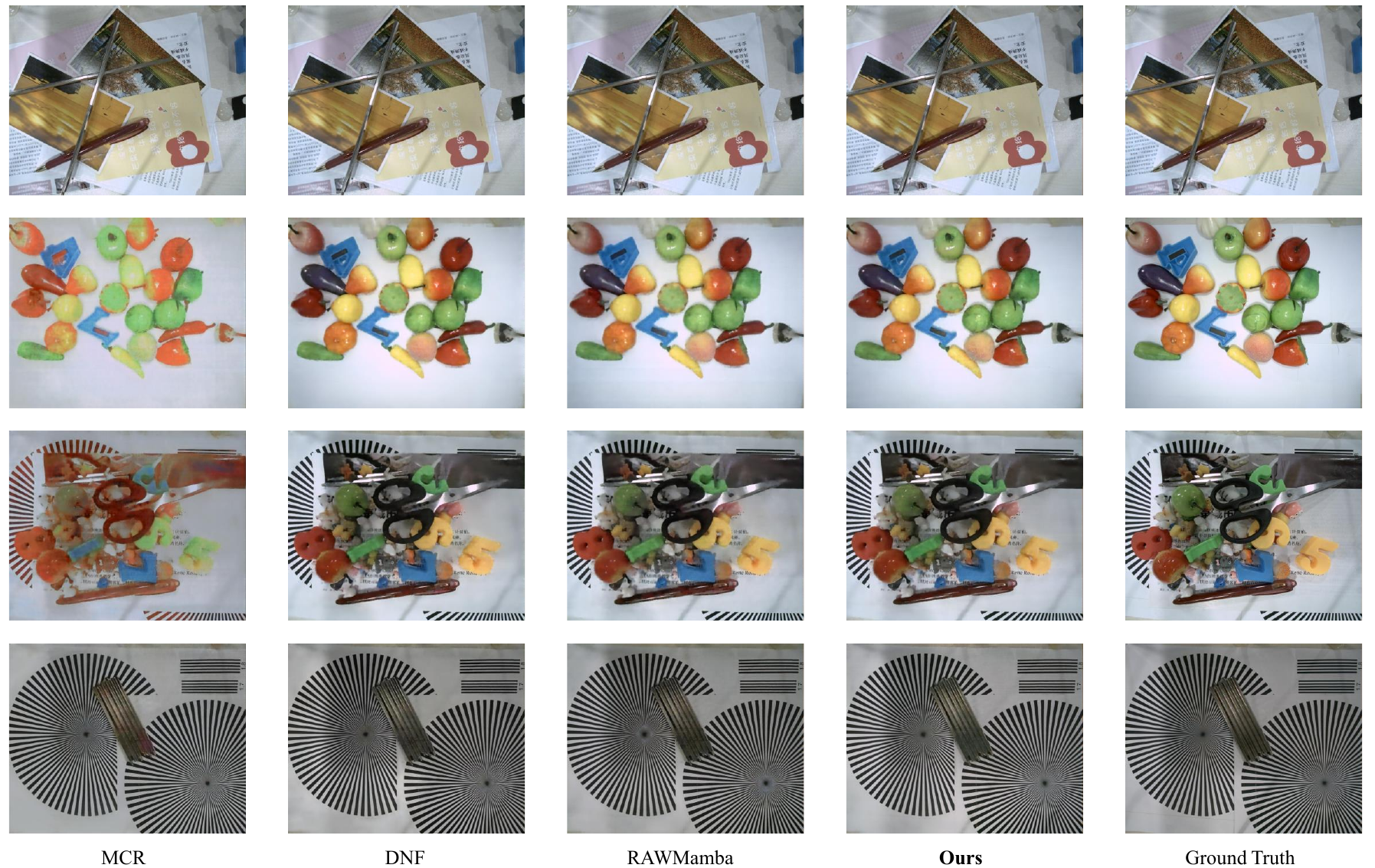}
    \caption{More visualization results between our method and the state-of-the-art methods on MCR dataset. (Zoom-in for best view).}
\end{figure*}

\end{document}